\documentclass[sigconf]{acmart}

\usepackage{booktabs} 
\usepackage{mathrsfs}
\usepackage{multirow}
\usepackage{graphicx}
\usepackage{epsfig}
\usepackage{graphics}
\usepackage{subfigure}
\usepackage{psfrag}
\usepackage{stfloats}
\usepackage{algorithm}
\usepackage{algorithmic}
\usepackage{stfloats}
\usepackage{amsmath}
\usepackage{amssymb}
\usepackage{amsfonts}
\usepackage{color}
\usepackage{booktabs}
\usepackage{slashbox}

\begin{document}

\copyrightyear{2017}
\acmYear{2017}
\setcopyright{acmcopyright}
\acmConference{SIGIR '17}{August 07-11, 2017}{Shinjuku, Tokyo, Japan}\acmPrice{15.00}\acmDOI{10.1145/3077136.3080842}
\acmISBN{978-1-4503-5022-8/17/08}

\fancyhead{}
\settopmatter{printacmref=false}

\title{Deep Semantic Hashing with\\ Generative Adversarial Networks}
\titlenote{{\small This work was performed at Microsoft Research Asia. The first two authors made equal contributions to this work.}}
\author{Zhaofan Qiu}
\affiliation{%
  \institution{University of Science and Technology of China}
  \city{Hefei}
  \country{China}
}
\email{zhaofanqiu@gmail.com}

\author{Yingwei Pan}
\affiliation{%
  \institution{University of Science and Technology of China}
  \city{Hefei}
  \country{China}
}
\email{panyw.ustc@gmail.com}

\author{Ting Yao}
\affiliation{%
  \institution{Microsoft Research Asia}
  \city{Beijing}
  \country{China}}
\email{tiyao@microsoft.com}

\author{Tao Mei}
\affiliation{%
  \institution{Microsoft Research Asia}
  \city{Beijing}
  \country{China}}
\email{tmei@microsoft.com}

\begin{abstract}
Hashing has been a widely-adopted technique for nearest neighbor search in large-scale image retrieval tasks. Recent research has shown that leveraging supervised information can lead to high quality hashing. However, the cost of annotating data is often an obstacle when applying supervised hashing to a new domain. Moreover, the results can suffer from the robustness problem as the data at training and test stage could come from similar but different distributions. This paper studies the exploration of generating synthetic data through semi-supervised generative adversarial networks (GANs), which leverages largely unlabeled and limited labeled training data to produce highly compelling data with intrinsic invariance and global coherence, for better understanding statistical structures of natural data. We demonstrate that the above two limitations can be well mitigated by applying the synthetic data for hashing. Specifically, a novel deep semantic hashing with GANs (DSH-GANs) is presented, which mainly consists of four components: a deep convolution neural networks (CNN) for learning image representations, an adversary stream to distinguish synthetic images from real ones, a hash stream for encoding image representations to hash codes and a classification stream. The whole architecture is trained end-to-end by jointly optimizing three losses, i.e., adversarial loss to correct label of synthetic or real for each sample, triplet ranking loss to preserve the relative similarity ordering in the input real-synthetic triplets and classification loss to classify each sample accurately. Extensive experiments conducted on both CIFAR-10 and NUS-WIDE image benchmarks validate the capability of exploiting synthetic images for hashing. Our framework also achieves superior results when compared to state-of-the-art deep hash models.
\end{abstract}

%
%



\maketitle

\section{Introduction}
Accelerated by tremendous increase in Internet bandwidth and storage space, multimedia data have been generated, published and spread explosively. This has led to the surge of research activities in large scale visual search. One fundamental research problem is similarity search, i.e., nearest neighbor search, which attempts to identify similar instances according to a query example. The need to search for millions of visual examples in a high-dimensional feature space, however, makes the task computationally expensive and thus very challenging.

Hashing techniques, one direction of the most well-known Approximate Nearest Neighbor (ANN) search methods, have been studied extensively due to its great efficiency in gigantic data. The basic idea of hashing is to construct a series of hash functions to map each example into a compact binary code, making the Hamming distances on similar examples minimized and simultaneously maximized on dissimilar examples. In the literature, there have been several techniques, including traditional hashing models based on hand-crafted features \cite{Gionis:VLDB99,Gong:PAMI13,Wang:PAMI12,Liu:CVPR12} and deep models \cite{Lai:CVPR15,Liong:CVPR15}, being proposed for addressing the problem of hashing. The former seek hashing function on hand-crafted features, which separate the encoding of feature representations and their quantization to hash codes, resulting in sub-optimal solution. The latter jointly learn feature representations and projections from them to hash codes in a deep architecture. While encouraging performances are reported in the aforementioned approaches especially when supervised information is available, we are often facing the problems of applying these methods to new applications where there is only few labeled training data, not to mention that the distribution of training data may be even different with that in test stage.

We demonstrate in this paper that the above limitations can be mitigated by generating synthetic data for training through Generative Adversarial Networks (GANs). GANs is a new recently proposed framework for estimating generative models via an adversarial process. The spirit behind is a minimax two-player game, in which a generative model is to capture the data distribution and a discriminative model aims to estimate the probability that a sample is from the real training data rather than the generative model. The generative model and discriminative model are trained simultaneously and the learning of the generative model is to fool the discriminative model into making mistakes. Once the training is complete, GANs is capable of generating both diverse and discriminable training examples, which have a great potential to characterize the statistical structures of natural data.

\begin{figure*}[!tb]
\centering {\includegraphics[width=0.99\textwidth]{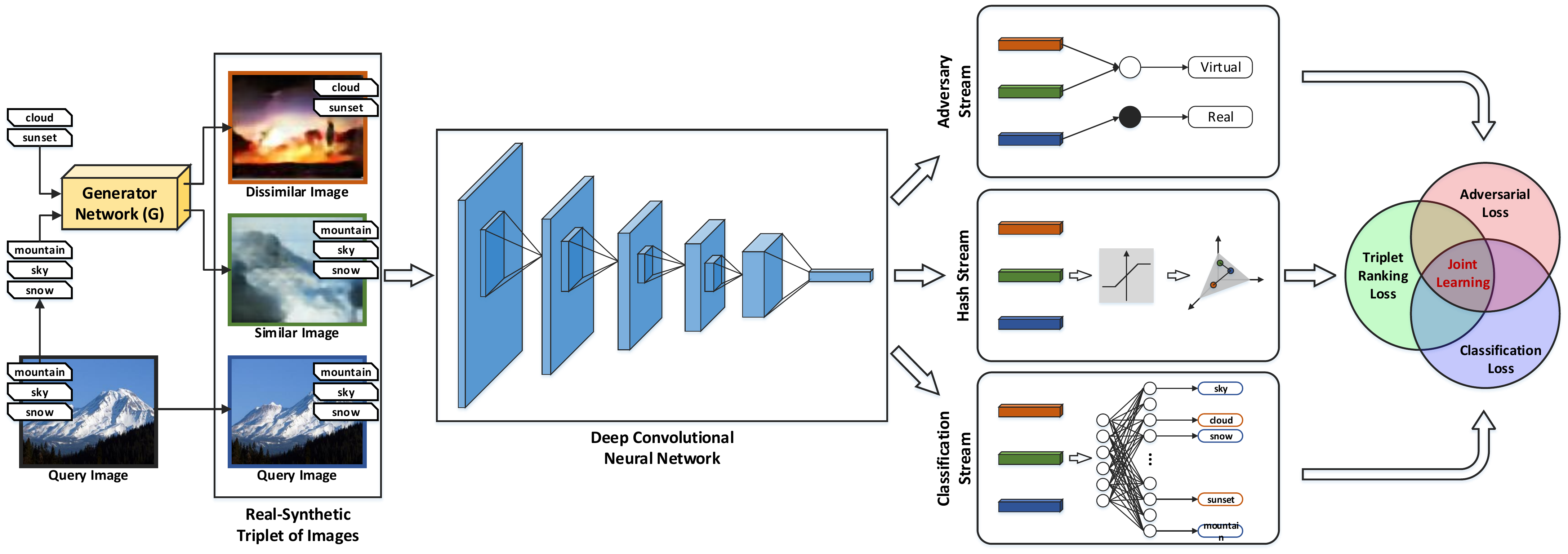}}
\caption{\small Deep Semantic Hashing with GANs (DSH-GANs) framework (better viewed in color). The input to DSH-GANs architecture is in the form of real-synthetic image triplets and each tuple consists of one real image as query image, one synthetic and similar image produced with same labels of query image through generator network $G$, and another synthetic but dissimilar image synthesized by $G$ conditioning on different labels. A shared deep convolutional neural networks is exploited for learning image representations, followed by three streams, i.e., hash stream, adversary stream and classification stream. Hash stream is to encode each image into hash codes with relative similarity preservation measured by a triplet ranking loss. Adversary stream is to distinguish synthetic images from real ones trained with an adversarial loss. Classification stream is to characterize the semantic structures on image and softmax loss or cross entropy loss is computed for single label and multi-label classification, respectively. The whole architecture is jointly optimized in an end-to-end fashion.}
\label{fig:fig1}
\end{figure*}

By consolidating the idea of generating training data for boosting hashing, we present a novel Deep Semantic Hashing with GANs (DSH-GANs) architecture, as shown in Figure \ref{fig:fig1}. Specifically, a semi-supervised GANs is first pre-trained on both labeled and unlabeled training data to produce synthetic examples conditioning on class labels. Then, we form a set of real-synthetic triplets and each tuple contains one real image as query image, one synthetic and semantically similar image and another synthetic but dissimilar image. A shared CNN is exploited to capture image representations, followed by importing into an adversary stream for differentiating the synthetic images from real ones, a hash stream to encode hash codes and a classification stream for measuring semantics. An adversarial loss is computed to correct the predicted labels (i.e., synthetic or real) of the images in adversary stream and a triplet ranking loss is devised to preserve relative similarities at the top of hash stream. Meanwhile, a classification error is formulated in classification stream. By jointly learning the three streams, our DSH-GANs is expected to offer a hashing model with high generalization ability and the generated hash codes could better reflect semantic relations between images. It is also worth noting that the whole architecture is trainable in an end-to-end fashion.

In summary, this paper makes the following contributions:

(1) We explore the problem of supervised hashing by exploiting the synthetic training data from GANs. To the best of our knowledge, this paper represents the first effort towards this target in the information retrieval research community.

(2) A novel hashing architecture, which combines adversary process, hash coding and classification, is proposed to enhance the generalization ability of hashing model and produce hash codes, which preserve not only relative similarity between images but also semantics of images.

(3) Extensive experiments on two widely used datasets demonstrate the advantages of our proposal over several state-of-the-art hashing techniques.

\section{Related Work}\label{sec:RW}
We briefly group the related works into two categories: hashing for image search, image synthesis with Generative Adversarial Networks (GANs). The former draws upon research in encoding visual images into compact binary codes for efficient image search, while the latter investigates synthesizing realistic images by utilizing GANs.

\textbf{Hashing for Image Search.}
The research in this direction has proceeded along two dimensions: hand-crafted features based hashing and deep architectures for hashing.

There are three main directions on hand-crafted feature based hashing: unsupervised hashing, semi-supervised hashing and supervised hashing. Unsupervised hashing~\cite{Gionis:VLDB99,Gong:PAMI13} refers to the setting when the label information is not available. Locality Sensitive Hashing (LSH) \cite{Gionis:VLDB99} is one of the most popular unsupervised hashing methods, which simply uses random linear projections to construct hash functions. This method is subsequently expanded to Kernelized and Multi-Kernel Locality Sensitive Hashing \cite{Kulis:PAMI12,Xia:SIGIR12}. Another effective method named Iterative Quantization (ITQ) \cite{Gong:PAMI13} is proposed for better quantization rather than random projections. Semi-supervised hashing approaches attempt to improve the quality of hash codes by leveraging supervised information into learning procedure. For example, Wang \emph{et al.} develop a Semi-Supervised Hashing (SSH) \cite{Wang:PAMI12} which utilizes pairwise information on labeled samples to preserve semantic similarity. In another work \cite{Kim:ICDM11}, Semi-Supervised Discriminant Hashing (SSDH) learns hash codes based on Fisher's discriminant analysis to maximize separability between labeled data from different classes while the unlabeled data are exploited for regularization. When all label information is available, we refer to the problem as supervised hashing. The representative in this category is Kernel-based Supervised Hashing (KSH) \cite{Liu:CVPR12} which utilizes pairwise relationship between examples to achieve high quality hashing.

Inspired by recent advances in visual representation learning \cite{Alex:NIPS12,qiu2017deep,pan2016learning} by using deep convolutional neural networks, several deep architecture based hashing methods have been proposed. Semantic Hashing \cite{Salakhutdinov:IJAR09} is one of the earlier works to exploit deep learning techniques for hashing. It applies the stacked Restricted Boltzman Machine (RBM) \cite{Hinton:SC06} to learn binary hash codes for visual search. Recently, Xia \emph{et al.} propose Convolutional Neural Networks Hashing (CNNH)~\cite{Xia:AAAI14} to decompose the hash learning process into a stage of learning approximate hash codes with the pairwise relationship and a following stage of simultaneously learning image feature and hash function. Later in~\cite{Li:IJCAI16}, such a two-stage method with pairwise labels is further developed into an end-to-end system, Deep Pairwise-Supervised Hashing (DPSH), which performs simultaneous feature learning and hash encoding. Similar in spirit, Network In Network Hashing (NINH) \cite{Lai:CVPR15} incorporates the supervised information among triplet labels into the feature learning based deep hashing architecture. More recently, Zhu \emph{et al.} devise Deep Hashing Network (DHN) to simultaneously optimize the pairwise cross-entropy loss on semantically similar pairs and the pairwise quantization loss on compact hash codes for hashing in \cite{zhu2016deep}.

In summary, our work belongs to deep architecture based hashing. The aforementioned deep approaches often focus on leveraging supervised information for training CNNs. Our work in this paper contributes by not only exploring image semantic supervision for hash learning, but also preserving relative similarity between real and synthetic images which are generated through a semi-supervised GANs with intrinsic invariance and global coherence.

\textbf{Image Synthesis with GANs.}
Synthesizing realistic images has been studied and analyzed widely in AI systems for characterizing the pixel level structure of natural images. Thanks to the recent development of Generative Adversarial Networks (GANs), researchers have strived to automatically synthesize image with GANs, which could be regarded as the generator network modules learnt with a two-player minimax game mechanism. Goodfellow \emph{et al.} propose a theoretical framework of GANs and utilize GANs to generate images without any supervised information in \cite{Goodfellow:NIPS14}. Although the earlier GANs offer a distinct and promising direction for image synthesis, the results are somewhat noisy and blurry. Hence, Laplacian pyramid is further incorporated into GANs in \cite{Denton:NIPS15} to produce high quality images. Later in \cite{Radford:ICLR16}, Radford \emph{et al.} devise deep convolutional generative adversarial networks (DCGANs) for unsupervised representation learning.

The aforementioned three works mainly explore image synthesis task in an unconditioned manner that generates synthetic images without any supervised information. Another direction of image synthesis with GANs is to synthesize images by conditioning on supervised information (e.g., class labels or text descriptions). \cite{Mirza2014conditional} is one of the earliest works that develop a conditional version of GANs by additionally feeding class labels into both discriminator and generator of GANs. Later in \cite{Odena2016conditional}, this model is further expended with a specialized cost function for classification, named auxiliary classifier GANs (AC-GANs), for generating synthetic images with global coherence and high diversity. Recently, Reed \emph{et al.} utilize GANs for image synthesis based on given text descriptions in \cite{Reed:ICML16}, enabling translation from character level to pixel level.

Most of the above approaches focus on leveraging GANs for image synthesis. Our work is different that we apply the synthetic images generated from GANs learnt on both largely unlabeled and limited labeled images for hash learning, leading to more effective and robust binary image representation for image retrieval task.

\begin{figure*}[!tb]
\centering {\includegraphics[width=0.92\textwidth]{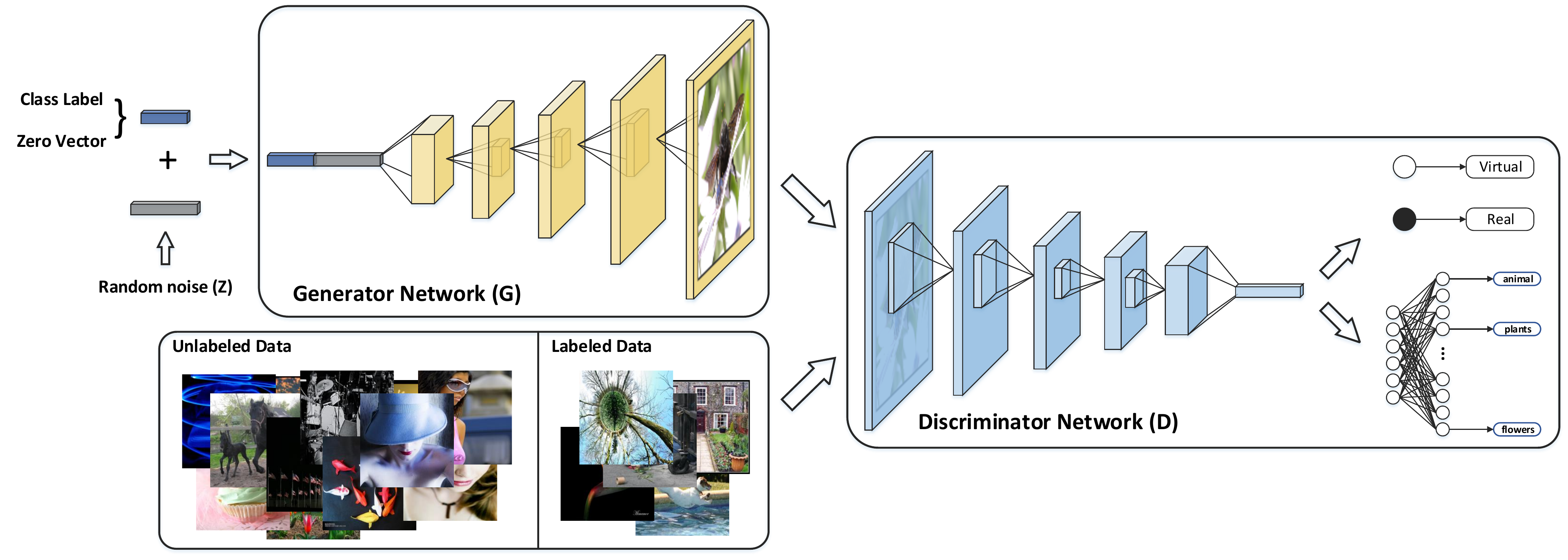}}
\caption{\small Our semi-supervised GANs framework mainly consists of a generator network $G$ and a discriminator network $D$ (better viewed in color). For the generator network $G$, it tries to synthesize realistic images with the concatenation input of the class label vector ${\bf{C}}$ and random noise vector ${\bf{z}}$. For the discriminator network $D$, it tries to simultaneously distinguish real images from synthetic ones and classify input images with correct class labels. The whole architecture is trained with the adversarial loss for assigning correct source and the classification loss for assigning correct class label in a two-player minimax game mechanism.}
\label{fig:fig2}
\vspace{-0.1in}
\end{figure*}

\section{Deep Semantic Hashing with GANs (DSH-GANs)}
In this section, we will present the proposed Deep Semantic Hashing with GANs (DSH-GANs) in detailcon. Figure \ref{fig:fig1} illustrates an overview of our architecture for hash learning, which consists of four components: a shared CNN for learning image representations, an adversary stream for distinguishing synthetic images from real ones, a hash stream for encoding each image into hash codes and a classification stream for leveraging semantic supervision. Specifically, a semi-supervised GANs is first devised to leverage both unlabeled and labeled images for producing synthetic images conditioning on class labels, followed by the three streams in our proposed DSH-GANs framework. In particular, hash stream is trained with the input real-synthetic triplets in a triplet-wise manner, adversary stream recognizes the label of synthetic or real for each image example while classification stream reinforces the hash learning to preserve semantic structures on both real and synthetic images. Finally, the whole optimization of DSH-GANs and hash codes generation for image retrieval are elaborated.

\subsection{Notation}
Suppose there are $n$ images in the whole set, represented as: $\mathcal {X} = \{ {x_i | i = 1, \cdots,n}\}$ and each image can be presented as $x$. Similarly, assume there are $L~(L < n)$ labeled images and the set of the labeled images are denoted as $\mathcal {X}_l = \{ {{x}_j | j = 1, \cdots,L}\}$. The goal of image hashing is to learn a mapping $\mathcal{H}:x\to\{0,1\}^{K}$, such that an input image $x$ will be encoded into a \emph{K}-bit binary code $\mathcal{H}(x)$.

\subsection{Semi-supervised GANs}\label{ssec:SG}
An unconditional generative adversarial networks (GANs) consists of two networks: a generator network $G$ that captures the data distribution for synthesizing image and a discriminator network $D$ that distinguishes real images from synthetic ones. In particular, the generator network $G$ takes a random noise vector ${\bf{z}}$ as input and produces a synthetic image ${x_{syn}} = G\left( {\bf{z}} \right)$. For the discriminator network $D$, it takes an image $x$ as input stochastically chosen (with equal probability) from training real images or synthetic images through $G$ and produces a probability distribution $P\left( {S|x} \right) = D\left( x \right)$ over the two image sources. As proposed in \cite{Goodfellow:NIPS14}, the whole GANs can be trained in a two-player minimax game. Concretely, given an image sample $x$, the discriminator network $D$ is trained to minimize the adversarial loss, i.e., maximizing the log-likelihood of assigning correct source to this sample:
\begin{equation}\label{Eq:Eq1}\small
l_a\left( x \right) = \left\{ \begin{array}{l}
-\log P\left( {S = real|x} \right),x \in \mathcal {X}\\
-\log P\left( {S = synthetic|x} \right),x \in \mathcal {X}_{syn}
\end{array} \right.,
\end{equation}
where $\mathcal {X}$ and $\mathcal {X}_{syn}$ denote the collections of real images in training and synthetic images produced by $G$, respectively. Meanwhile, the generator network $G$ is trained to maximize the adversarial loss in Eq.(\ref{Eq:Eq1}), targeting for maximally fooling the discriminator network $D$ with its generated synthetic images $\mathcal {X}_{syn}$.

To characterize the pixel level structure of both unlabeled and labeled natural images in one architecture elegantly, we take the inspiration from conditional GANs \cite{Mirza2014conditional,Odena2016conditional} purely trained with supervised samples and devise a novel semi-supervised GANs architecture as shown in Figure \ref{fig:fig2}. Similar to aforementioned architectures of unconditional GANs, our semi-supervised GANs consists of a generator network $G$ for synthesizing images conditioning on class labels, and a discriminator network $D$ that simultaneously distinguishes real images from synthetic ones and classify the input images with correct class labels. Specifically, given the whole image set $\mathcal {X}$ including $L$ labeled images in $c$ classes, the class label information of each labeled image is first encoded into a $c$-dimensional vector ${\bf{C}} \in \{0,1\}^{c}$, whose element is a class label indicator. The indicator is 1 if the image contains this label otherwise the indictor is 0. As such, the class label vector ${\bf{C}}$ of each unlabeled image is set as zero vector ${\bf{0}}$. Then the generator network $G$ takes the concatenation of the class label vector ${\bf{C}}$ and random noise vector ${\bf{z}}\in \mathcal{N}(0,1)$ as input for producing a synthetic image ${x_{syn}} = G\left({\bf{C}}, {\bf{z}} \right)$. The discriminator network $D$ generates both a probability distribution over two sources and a probability distribution over all the $c$ class labels, i.e., $\left\{ {P\left( {S|x} \right),P\left( {C|x} \right)} \right\} = D\left( x \right)$, for each image example $x$ from either real images or synthetic images through $G$. It is worth noting that both the unlabeled and labeled images are included in the real image selection pool of $D$ for better understanding the statistical structures of natural data.

The overall objective function of our semi-supervised GANs is composed of two parts: the adversarial loss $l_a\left( x \right)$ in Eq.(\ref{Eq:Eq1}) for assigning correct source to the image example $x$, and the classification loss $l_c\left( x \right)$ for assigning correct class label to this image. The details of how to measure the classification loss for images with single label or multiple labels will be presented in Section \ref{ssec:CS}. Accordingly, the discriminator network $D$ is learnt to minimize $l_c\left( x \right) + l_a\left( x \right)$ for recognizing both correct source and class label, while the generator network $G$ is trained to minimize $l_c\left( x \right) - l_a\left( x \right)$ for fooling $D$ on source prediction and meanwhile preserving the correct class label. After training the whole semi-supervised GANs with unlabeled and labeled natural images, the learnt generator network $G$ is directly utilized as the pre-trained generator network in our DSH-GANs architecture for synthesizing realistic images conditioning on class labels.

\subsection{Hash Stream}\label{ssec:HS}
In the traditional binary representation learning, the hash encoding of each image is always treated independently in point-wise hashing learning methods, regardless of the relationships of similar or dissimilar between images. More importantly, the relative similarity relations like ``for query image $x$, it should be more similar to image $x^+$ than to image $x^-$," are reflected in the image class labels in view that image $x$ and $x^+$ belong to the same class while image $x^-$ comes from other categories. The utilization of these relative similarity relations has also been proved to be effective in hash coding~\cite{Lai:CVPR15,Pan2015semi,dai2016binary,yao2016deep}. Inspired by the idea of preserving relative similarity in deep architecture~\cite{Lai:CVPR15}, we propose a hash stream for encoding hash codes learnt in a triplet-wise manner, which aims to preserve the relative similarity ordering in the input real-synthetic triplets.

Specifically, we can easily obtain a set of real-synthetic triplets \mbox{$\mathcal{T}$} based on image labels, where each tuple $(x, x^{+}_{syn}, x^{-}_{syn})$ consists of one real image $x$ as query image, one synthetic and semantically similar image $x^{+}_{syn}$, and another synthetic but dissimilar image $x^{-}_{syn}$. Note that $x^{+}_{syn}$ is synthesized by generator network $G$ conditioning on the same class labels of query image $x$, while $x^{-}_{syn}$ is produced through $G$ conditioning on different labels of $x$. To preserve the similarity relations in the real-synthetic triplets, we aim to learn a hash mapping $\mathcal{H}(\cdot)$ which makes the compact code $\mathcal{H}(x)$ more similar to $\mathcal{H}(x^{+}_{syn})$ than to $\mathcal{H}(x^{-}_{syn})$. Hence, the triplet ranking loss is employed and defined as
\begin{equation}\label{Eq:Eq2}\small
\begin{split}
&\quad{{{\hat{l}}}_{triplet}}(x,{x^{+}_{syn}},{x^{-}_{syn}}) \\
&= \max (0,1-{{\left\| \mathcal{H}(x)-\mathcal{H}({x^{-}_{syn}}) \right\|}_{H}}+{{\left\| \mathcal{H}(x)-\mathcal{H}({x^{+}_{syn}}) \right\|}_{H}}) \\
&\quad s.t. \quad\mathcal{H}(x), \mathcal{H}(x^{+}_{syn}), \mathcal{H}(x^{-}_{syn})\in\{0,1\}^{K}
\end{split}~~,
\end{equation}
where $||\cdot||_H$ represents Hamming distance. For ease of optimization, natural relaxation tricks are utilized on Eq.(\ref{Eq:Eq2}) to change integer constraint to the range constraint and replace Hamming norm with $l_2$ norm. Then, the triplet ranking loss function is reformulated as
\begin{equation}\label{Eq:Eq3}\small
\begin{split}
&\quad{{{\hat{l}}}_{triplet}}(x,{x^{+}_{syn}},{x^{-}_{syn}}) \\
&= \max (0,1-{{\left\| \mathcal{H}(x)-\mathcal{H}(x^{-}_{syn}) \right\|}_{2}^{2}}+{{\left\| \mathcal{H}(x)-\mathcal{H}(x^{+}_{syn}) \right\|}_{2}^{2}}) \\
&\quad s.t. \quad\mathcal{H}(x), \mathcal{H}(x^{+}_{syn}), \mathcal{H}(x^{-}_{syn})\in[0,1]^{K}
\end{split}~~.
\end{equation}

\subsection{Adversary Stream}\label{ssec:AS}
Noticing that the input real-synthetic triplets of aforementioned hash stream contain not only different semantics, but also are from distinctly different sources. As a result, we additionally devise an adversary stream to distinguish synthetic images from real ones within each real-synthetic triplet, targeting for exploiting the mutual but also fuzzy relationship between the hash codes learning and source discrimination in GANs. In particular, for the adversary stream, the shared CNN for learning image representation can be treated as the discriminator network $D$ in GANs, followed by a cross entropy loss layer for source prediction. Thus, given the real-synthetic triplet $(x, x^{+}_{syn}, x^{-}_{syn})$, an adversarial loss is used to measure the correctness of the predicted source (i.e., real or synthetic) of all the three images:
\begin{equation}\label{Eq:Eq4}\small
{{\hat l}_a}(x,x_{syn}^ + ,x_{syn}^ - ) = \frac{1}{3}\left( {{l_a}\left( x \right) + {l_a}\left( {x_{syn}^ + } \right) + {l_a}\left( {x_{syn}^ - } \right)} \right),
\end{equation}
where $l_a(\cdot)$ denotes the log-likelihood adversarial loss for each image as in Eq.(\ref{Eq:Eq1}).

\subsection{Classification Stream}\label{ssec:CS}
Image labels not only provide knowledge in classification but also are useful supervised information for mining semantic structures in images. A valid question is how to leverage the semantic supervision into both hashing and GANs, and make the generated hash codes better reflecting semantic similarities between images. Hence, we propose a joint learning mechanism by combining hash stream, adversary stream and classification stream. In the classification stream, a classification error is measured based on the input real-synthetic triplets. Specifically, for the single label classification, we use softmax optimization method. Given an input image $x$, the softmax loss is then formulated as
\begin{equation}\label{Eq:Eq5}\small
{l_c}(x) =  - \sum\limits_{j = 1}^c {{I_{(y = j)}}log\frac{{{e^{{\boldsymbol{\theta }}_j^\top{\bf{x}}}}}}{{\sum\nolimits_{l = 1}^c {{e^{{\boldsymbol{\theta }}_l^\top{\bf{x}}}}} }}},
\end{equation}
where ${\bf{x}}$ is the output image representation of shared CNN for image $x$, ${\boldsymbol{\theta }}_j$ denotes the parameter matrix in a softmax layer and $y \in \{1,2,...,c\}$ represents image class label. The indicator function $I_\emph{condition}=1$ if $\emph{condition}$ is true; otherwise $I_\emph{condition}=0$.

If an image contains multiple class labels, we refer to this problem as multi-label classification. Cross entropy loss is then employed in this case. Similar to softmax loss, cross entropy loss is computed by
\begin{equation}\label{Eq:Eq6}\small
\begin{split}
  & {l_c}(x) =  - \sum\limits_{j = 1}^c \big[{{I_{({{\bf{C}}_j} = 1)}}\log \left( {P({{\bf{C}}_j} = 1|{\bf{x}})} \right)} \\
  & \quad\quad\quad\quad\quad + (1 - {I_{({{\bf{C}}_j} = 1)}})\log\left( {1 - P({{\bf{C}}_j} = 1|{\bf{x}})} \right)\big]\\
  & P({{\bf{C}}_j} = 1|{\bf{x}}) = \frac{1}{{1 + {e^{ - \boldsymbol{\delta} _j^\top{\bf{x}}}}}}
\end{split},
\end{equation}
where ${{\bf{C}}_j}$ denotes the $j$-th element in class label vector ${\bf{C}}$ and ${\boldsymbol{\delta }}_j$ denotes the parameter matrix in a sigmoid layer.

Hence, given the real-synthetic triplet $(x, x^{+}_{syn}, x^{-}_{syn})$, the classification error is calculated on all the three examples by
\begin{equation}\label{Eq:Eq7}\small
{{\hat l}_c}(x,x_{syn}^ + ,x_{syn}^ - ) = \frac{1}{3}\left( {{l_c}\left( x \right) + {l_c}\left( {x_{syn}^ + } \right) + {l_c}\left( {x_{syn}^ - } \right)} \right).
\end{equation}

\subsection{Optimization}\label{ssec:OP}
The overall training objective of DSH-GANs integrates the triplet ranking loss in Eq.(\ref{Eq:Eq3}), adversarial loss in Eq.(\ref{Eq:Eq4}) and classification error in Eq.(\ref{Eq:Eq7}). As our DSH-GANs is a variant of GANs architecture which mainly consists of generator network $G$ for image synthesis with labels and the shared CNN for image representation learning, we train the whole architecture in a two-player minimax game mechanism. In particular, for shared CNN in hash stream, we update its parameters according to the following overall loss:
\begin{equation}\label{Eq:Eq8}\small
\begin{array}{l}
{{\hat l}_{CNN}} = \sum\limits_{\mathcal{T}} \big[{{{\hat l}_{triplet}}(x,x_{syn}^ + ,x_{syn}^ - )} \\
\quad\quad\quad\quad\quad\quad + {{\hat l}_a}(x,x_{syn}^ + ,x_{syn}^ - ) + {{\hat l}_c}(x,x_{syn}^ + ,x_{syn}^ - )\big]
\end{array},
\end{equation}
where $\mathcal{T}$ is the set of real-synthetic triplets. By minimizing this term, the shared CNN in hash stream is trained to preserve the relative similarity ordering in the real-synthetic triplets and simultaneously recognize both correct sources and class labels of images in the triplets.

For the generator network $G$, its parameters are adjusted with the following loss:
\begin{equation}\label{Eq:Eq9}\small
\begin{array}{l}
~~~~~~~~{{\hat l}_{G}} = \sum\limits_{\mathcal{T}} \big[{{{\hat l}_{triplet}}(x,x_{syn}^ + ,x_{syn}^ - )} \\
\quad\quad\quad\quad\quad - {{\hat l}_a}(x,x_{syn}^ + ,x_{syn}^ - ) + {{\hat l}_c}(x,x_{syn}^ + ,x_{syn}^ - )\big]
\end{array}.
\end{equation}
Thus, the generator network $G$ is trained to fool the shared CNN on source prediction and meanwhile preserve the relative similarity ordering and correct class labels of the real-synthetic triplets.

\subsection{Image Retrieval}\label{ssec:IR}
After the optimization of DSH-GANs, we can employ hash stream in the architecture followed by a sigmoid layer to generate \emph{K}-bit hash codes for each input image. In this procedure, an image $x$ is first encoded into a \emph{K}-dimension feature vector $\mathbf{h} = \mathcal{H}(x)$. Then, a quantization operation $\mathbf{b}=\mathcal{Q}(\mathbf{h})$ is exploited to generate hash codes $\mathbf{b}$, where $\mathcal{Q}(\mathbf{h})$ is a sign function on vector $\mathbf{h}$ with $\mathcal{Q}(h_i)=1$ if $h_i>0.5$ and otherwise $\mathcal{Q}(h_i)=0$. Given a query image, the retrieval list of images is produced by sorting the Hamming distances of hash codes between the query image and images in search~pool.

\section{Experiments}
We conducted extensive evaluations of our proposed architecture on two image datasets, i.e., CIFAR-10\footnote{http://www.cs.toronto.edu/~kriz/cifar.html} which is a collection of tiny images and NUS-WIDE\footnote{http://lms.comp.nus.edu.sg/research/NUS-WIDE.htm} of a large-scale Web image dataset.

\subsection{Datasets}
The \textbf{CIFAR-10} dataset consists of 60,000 real world tiny images (32$\times$32 pixels), which can be divided into 10 categories and 6,000 images for each category. We randomly select 1,000 images (100 images per class) as the test query set. For the unsupervised setting, all the rest images are used as training samples. For the supervised setting, we additionally sample 500 images from each class in the training samples and constitute a subset of 5,000 labeled images for training. The rest training images are treated as the unlabeled data.

The \textbf{NUS-WIDE} dataset contains 269,648 images collected from Flickr. Each of these images is associated with one or multiple labels in 81 semantic concepts. For a fair comparison, we follow the settings in~\cite{Lai:CVPR15} to employ the subset of images associated with 21 most frequent labels, where each label associates with at least 5,000 images. Similar to the split in CIFAR-10, we randomly select 2,100 images (100 images per class) as the test query set. For the unsupervised setting, all the rest images are used as the training set. For the supervised setting, we uniformly sample 500 images from each class to construct the labeled subset for training and the rest training images are all treated as unlabeled data.

\subsection{Experimental Settings}
On both datasets, we utilize AlexNet~\cite{Alex:NIPS12} as our basic CNN architecture and take the outputs of fc6 layer from AlexNet as the image representation. The shared CNN architecture is pre-trained on ImageNet dataset \citep{ILSVRC15} and the generator network $G$ is pre-trained with our proposed semi-supervised GANs on each dataset.

We mainly implement our proposed method based on Caffe \cite{Jia:MM14}, which is one of the widely adopted deep learning frameworks. For the semi-supervised GANs, we follow the standard settings in \cite{Radford:ICLR16} and train our GANs models on both datasets by utilizing Adam optimizer with a mini-batch size of 128. All weights are initialized from a zero-centered Normal distribution with standard deviation 0.02 and the slope of the leak is set to 0.2 in the LeakyReLU. We fix the learning rate and momentum to $0.0002$ and $0.9$, respectively. For our DSH-GANs architecture, it is trained by stochastic gradient descent with $0.9$ momentum. The start learning rate is set to $0.0001$, and we decrease it to 10\% after $10,000$ iterations on CIFAR-10 and after $40,000$ iterations on NUS-WIDE, respectively. The mini-batch size of images is $64$ and the weight decay parameter is $0.0005$.

\subsection{Protocols and Baseline Methods}
We follow four evaluation protocols, i.e., mean average precision (MAP), hash lookup, precision-recall curve, and precision curves w.r.t. different numbers of top returned samples, which are widely used in \cite{Gong:PAMI13,Liu:CVPR12,Lai:CVPR15}. We compare the following approaches for performance evaluation:

(1) Locality Sensitive Hashing \cite{Gionis:VLDB99} (LSH) aims to map similar examples to the same bucket with high probability by using a Gaussian random projection matrix. The property of locality in the original space will be largely preserved in the Hamming space.

(2) Spectral Hashing \cite{Weiss:NIPS08} (SH) is based on quantizing the values of analytical eigenfunctions computed along PCA directions of the data.

(3) Iterative Quantization \cite{Gong:PAMI13} (ITQ) learns similarity-preserving binary codes by directly minimizing the quantization error of mapping data to vertices of the binary hypercube.

(4) Kernel-based Supervised Hashing \cite{Liu:CVPR12} (KSH) employs a kernel formulation for learning the hash functions to handle linearly inseparable data.

(5) Convolutional Neural Networks Hashing \cite{Xia:AAAI14} (CNNH) firstly learns approximate hash codes with the supervised pairwise relationship and then trains CNN architecture with approximate hash codes and image tags.

(6) Network In Network Hashing \cite{Lai:CVPR15} (NINH) utilizes a triplet ranking loss to preserve relative similarity and divide-and-encode modules to encode hash bits.

(7) Deep Pairwise-Supervised Hashing \cite{Li:IJCAI16} (DPSH) performs simultaneous feature learning and hash learning by leveraging pairwise labels in an end-to-end system.

(8) Deep Semantic Hashing with Generative Adversarial Networks (DSH-GANs) is our proposal in this paper. A slightly different of this run is named as DSH-GANs$^{-}$, which is trained without classification.

Note that for the four hashing methods using hand-crafted features (i.e., LSH, SH, ITQ and KSH), each image in CIFAR-10 and NUS-WIDE is represented by a 512-dimensional GIST vector and an officially available 500-dimensional bag-of-words vector, respectively. For the deep hashing methods, we resize all images to be 224$\times$224 pixels and then directly exploit the raw image pixels as input. Moreover, we also conduct the experiments by using the outputs of fc6 layer in AlexNet as image representation in the four traditional hashing approaches and name them as LSH+CNN, SH+CNN, ITQ+CNN and KSH+CNN, respectively.

\begin{table*}
\centering
\small
\caption{\small Accuracy in terms of MAP. The best MAPs for each category are shown in boldface. Note that the MAP performance is calculated on the top 5,000 returned images for NUS-WIDE dataset.}
\begin{tabular}{|l|*{4}{c|}*{3}{c|}c|}\hline
Method & \multicolumn{4}{c|}{CIFAR-10 (MAP)} & \multicolumn{4}{c|}{NUS-WIDE (MAP)} \\ \hline \hline
& {~12-bits~} & {~24-bits~} & {~32-bits~} & {~48-bits~} & {~12-bits~} & {~24-bits~} & {~32-bits~} & {~48-bits~} \\ \hline
DSH-GANs & \textbf{0.735} & \textbf{0.781} & \textbf{0.787} & \textbf{0.802} & \textbf{0.838} & \textbf{0.856} & \textbf{0.861} & \textbf{0.863} \\ \hline
DSH-GANs$^{-}$ & 0.726 & 0.769 & 0.772 & 0.783 & 0.823 & 0.847 & 0.845 & 0.854 \\ \hline \hline
DPSH & 0.713 & 0.727 & 0.744 & 0.757 & 0.794 & 0.822 & 0.838 & 0.851 \\ \hline
NINH & 0.552 & 0.566 & 0.558 & 0.581 & 0.674 & 0.697 & 0.713 & 0.715 \\ \hline
CNNH & 0.439 & 0.476 & 0.472 & 0.489 & 0.611 & 0.618 & 0.625 & 0.608 \\ \hline \hline
KSH+CNN & 0.446 & 0.502 & 0.518 & 0.516 & 0.746 & 0.774 & 0.765 & 0.749\\ \hline
ITQ+CNN & 0.212 & 0.230 & 0.234 & 0.240 & 0.728 & 0.707 & 0.689 & 0.661\\ \hline
SH+CNN & 0.158 & 0.157 & 0.154 & 0.151 & 0.620 & 0.611 & 0.620 & 0.591\\ \hline
LSH+CNN & 0.134 & 0.157 & 0.173 & 0.185 & 0.438 & 0.586 & 0.571 & 0.507\\ \hline
KSH & 0.303 & 0.337 & 0.346 & 0.356 & 0.556 & 0.572 & 0.581 & 0.588 \\ \hline
ITQ & 0.162 & 0.169 & 0.172 & 0.175 & 0.452 & 0.468 & 0.472 & 0.477\\ \hline
SH & 0.127 & 0.128 & 0.126 & 0.129 & 0.454 & 0.406 & 0.405 & 0.400\\ \hline
LSH & 0.121 & 0.126 & 0.120 & 0.120 & 0.403 & 0.421 & 0.426 & 0.441\\ \hline
\end{tabular}
\label{tab:tab1}

\end{table*}

\begin{figure*}[!tb]
   \centering
   \subfigure[]{
     \label{fig:fig5:a}
     \includegraphics[width=0.32\textwidth]{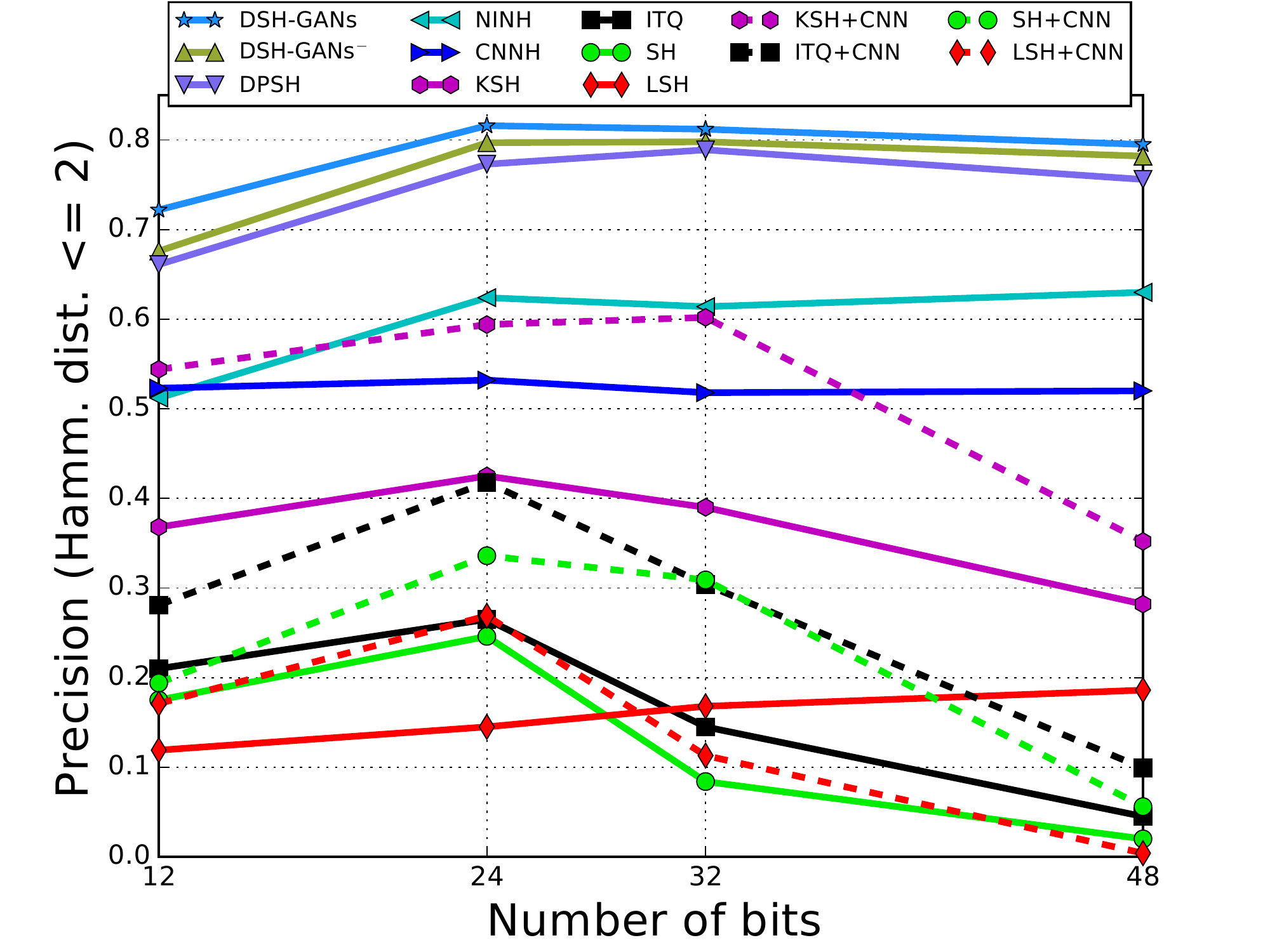}}
   \subfigure[]{
     \label{fig:fig5:b}
     \includegraphics[width=0.32\textwidth]{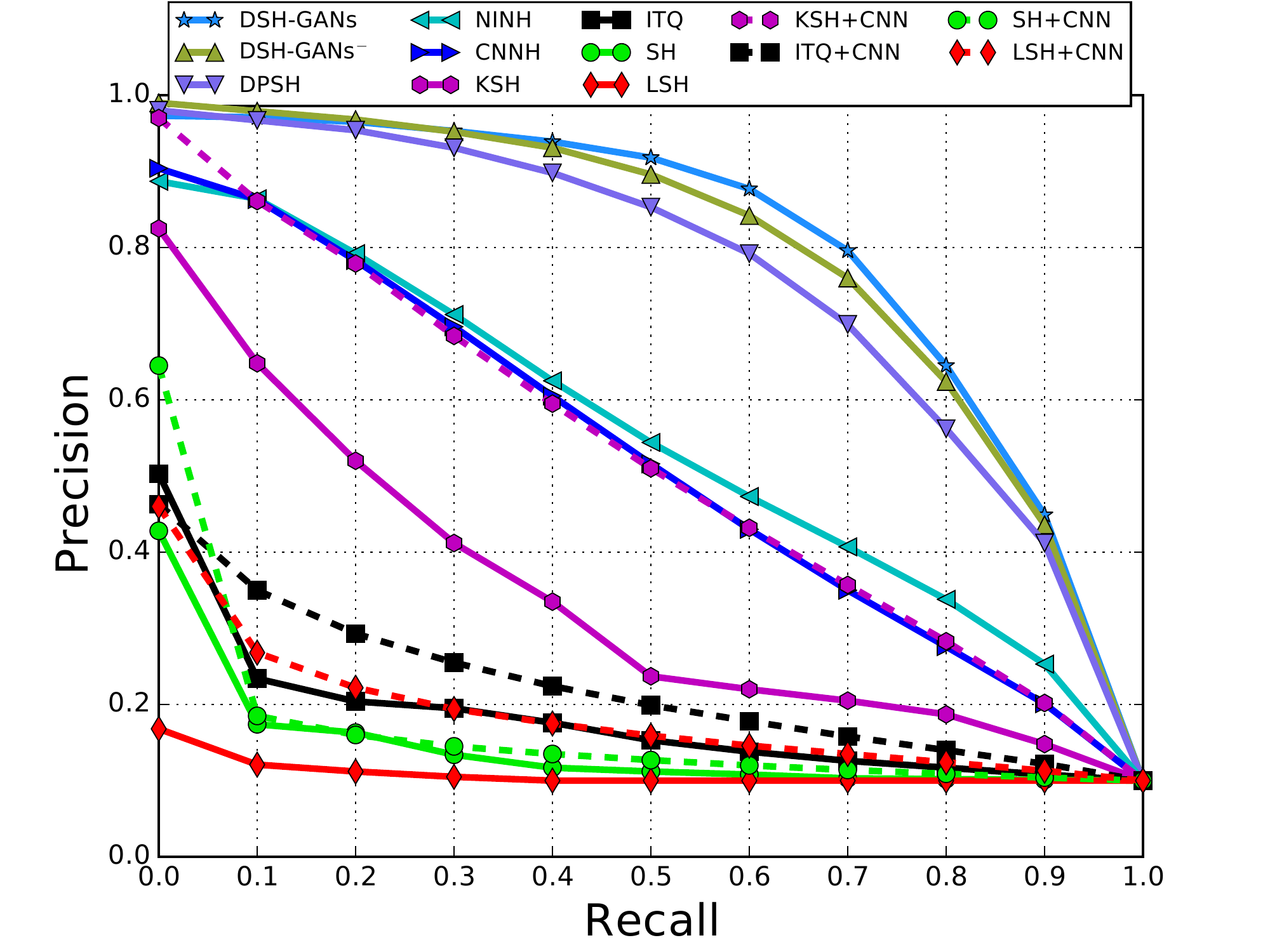}}
   \subfigure[]{
     \label{fig:fig5:c}
     \includegraphics[width=0.32\textwidth]{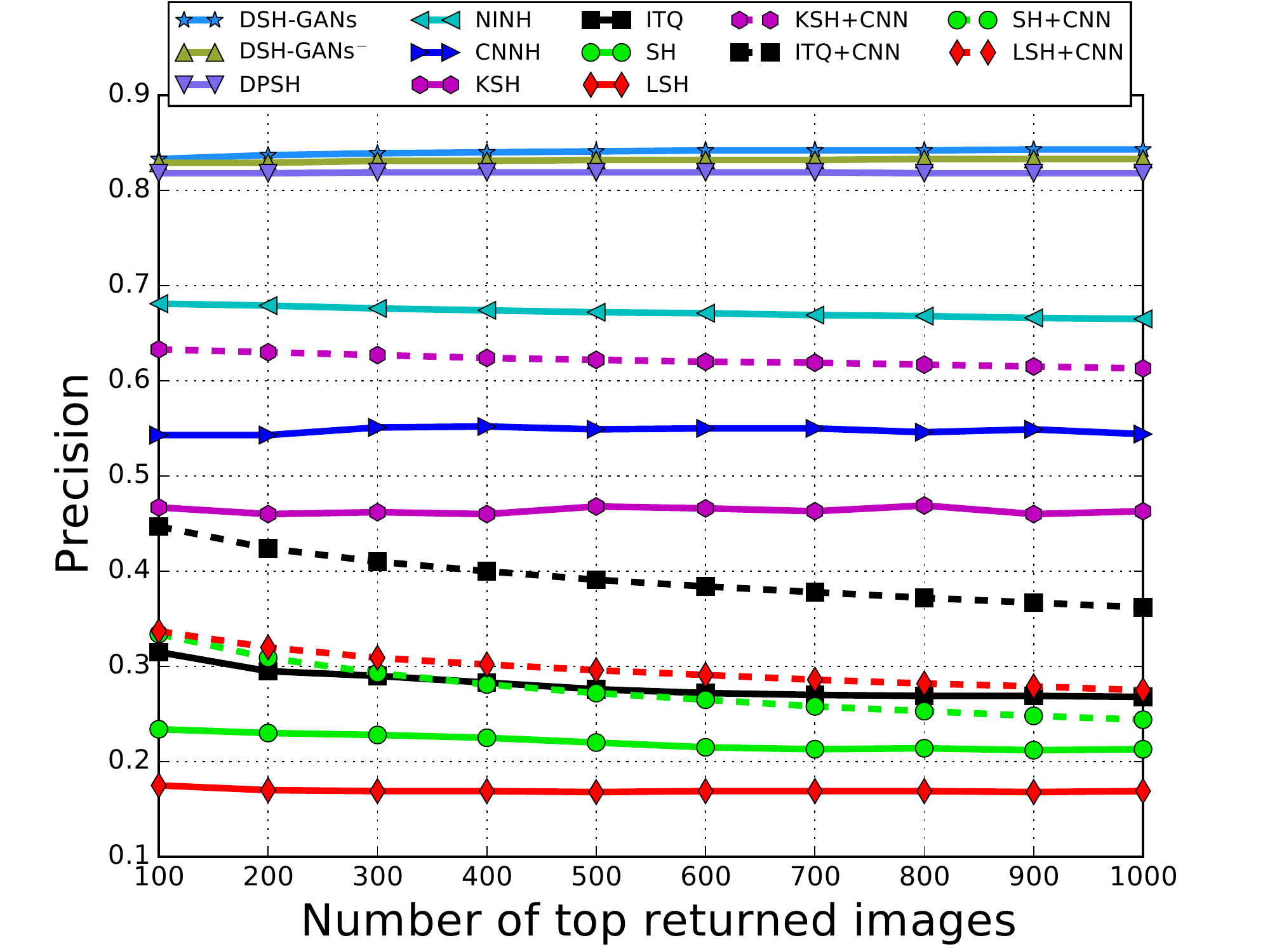}}
     \vspace{-0.08in}
   \caption{\small Comparisons with state-of-the-art approaches on CIFAR-10 dataset. (a) Precision within Hamming radius 2 using hash lookup. (b) Precision-Recall curves with 48-bits. (c) precision curves with 48-bits w.r.t. different number of top returned samples. Better viewed in original color pdf file.}
   \label{fig:fig5}
\end{figure*}

\subsection{Results on CIFAR-10 Dataset}
The left half of Table \ref{tab:tab1} shows the MAP performance comparisons on CIFAR-10 dataset. Overall, the results across different number of hash bits indicate that our DSH-GANs consistently outperforms others. In particular, the MAP of DSH-GANs with 48-bits makes the relative improvement over the best traditional competitor KSH with GIST features or the outputs of fc6 layer in AlexNet, and deep model DPSH by 125.3\%, 55.4\% and 5.9\%, respectively. Furthermore, traditional approaches with image representations extracted from CNN architecture lead to a large performance boost against these methods with GIST features, which is expected as deep CNN has demonstrated its high capability in generating image representations. Compared to the traditional models with deep image representations, deep hash models which benefit from the joint learning of image representations and hash coding exhibit better performances. DSH-GANs$^-$ outperforms DPSH and NINH. The result basically indicates the advantage of exploring synthetic images in hashing. DSH-GANs further improves DSH-GANs$^-$ with a relative increase of 1.2\%$\sim$2.4\%, demonstrating the strength of boosting hashing by additionally preserving semantics of images through~classification. In addition, when utilizing a deeper CNN architecture VGG-19 \cite{Simonyan:ICLR15} networks as our basic CNN, the MAP performance of our DSH-GANs with 12-bits, 24-bits, 32-bits and 48bits will be boosted up to 86.1\%, 88.1\%, 87.9\% and 88.4\%, respectively.

In the evaluation of hash lookup within Hamming radius 2 as shown in Figure \ref{fig:fig5:a}, the precisions for most of the traditional methods drop when a longer size of hash codes is used (48 bits in our case). This is because the number of samples falling into a bucket decreases exponentially for longer sizes of hash codes. Therefore, for some query images, there are not even any neighbor in a Hamming ball of radius 2. Even in this case, the precision of our proposed DSH-GANs only has a slight decrease from 80.6\% of 32 bits to 79.7\% of 48 bits, indicating fewer failed queries for DSH-GANs. We further detail the precision-recall curves and precision curves with 48-bits w.r.t. different number of top returned samples in Figure \ref{fig:fig5:b} and \ref{fig:fig5:c}. The results confirm the trends observed in Figure \ref{fig:fig5:a} and demonstrate performance improvements by our proposed DSH-GANs approach over other methods.

\begin{figure*}[!tb]
   \centering
   \subfigure[]{
     \label{fig:fig6:a}
     \includegraphics[width=0.32\textwidth]{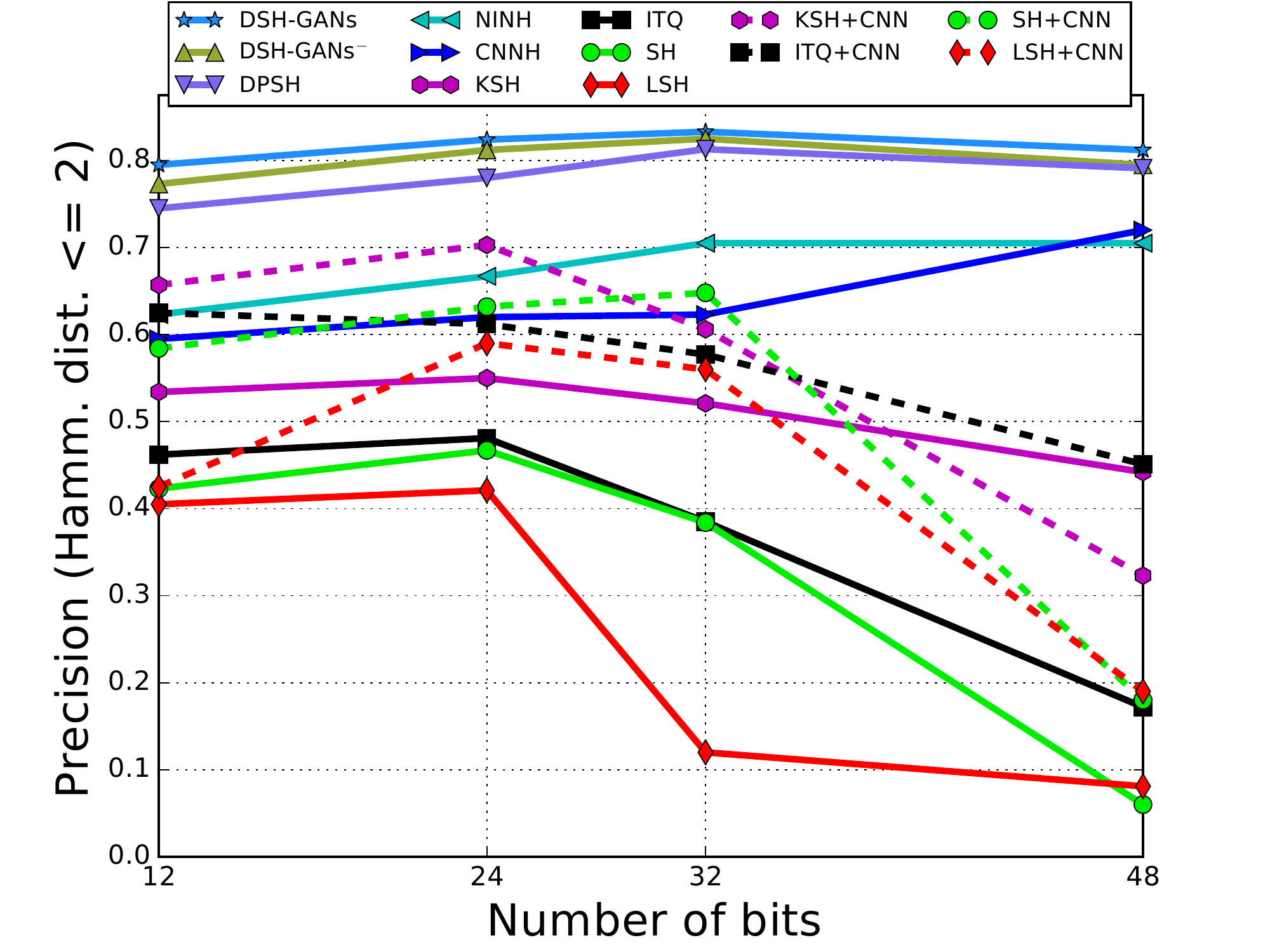}}
   \subfigure[]{
     \label{fig:fig6:b}
     \includegraphics[width=0.32\textwidth]{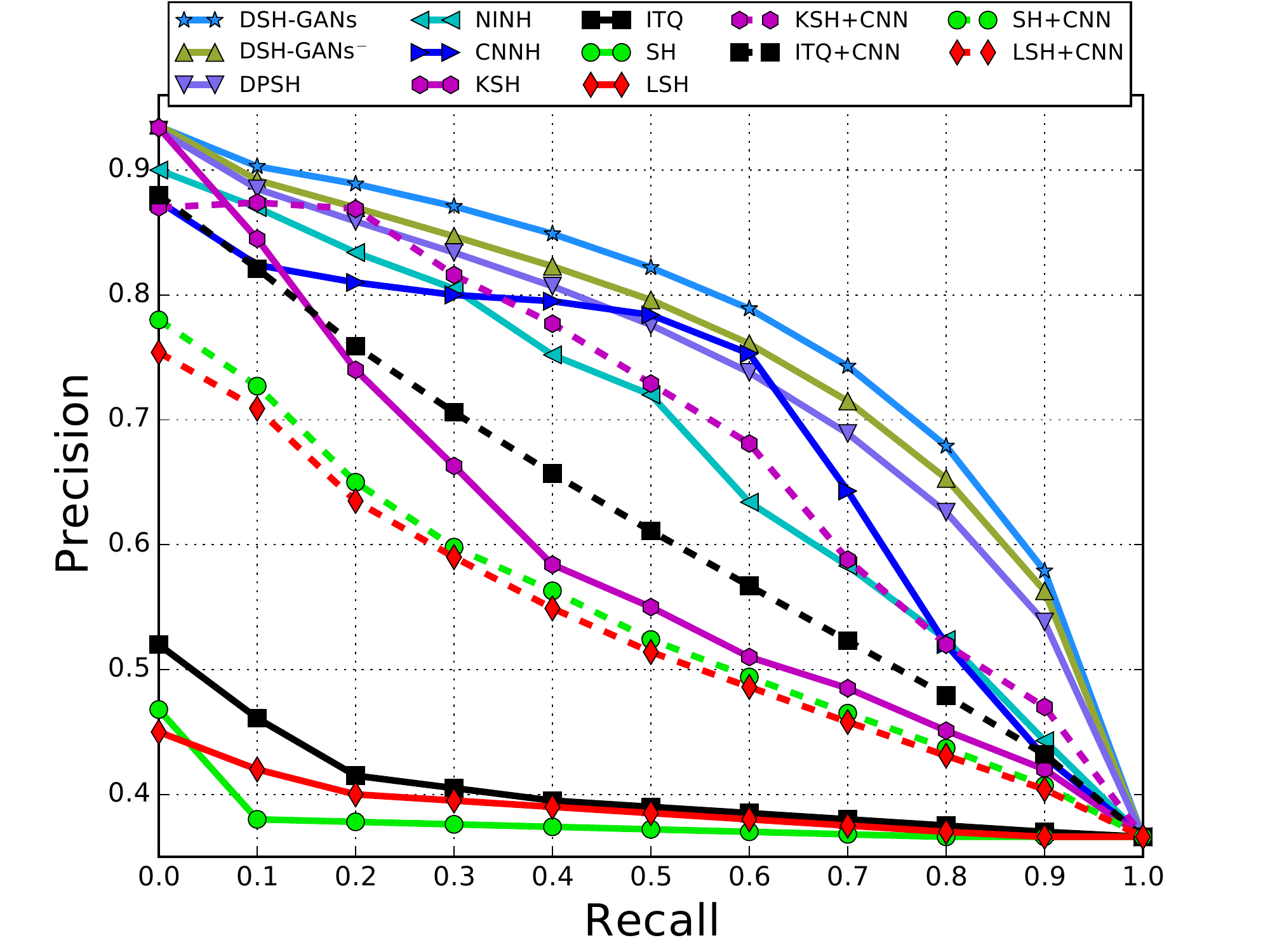}}
   \subfigure[]{
     \label{fig:fig6:c}
     \includegraphics[width=0.32\textwidth]{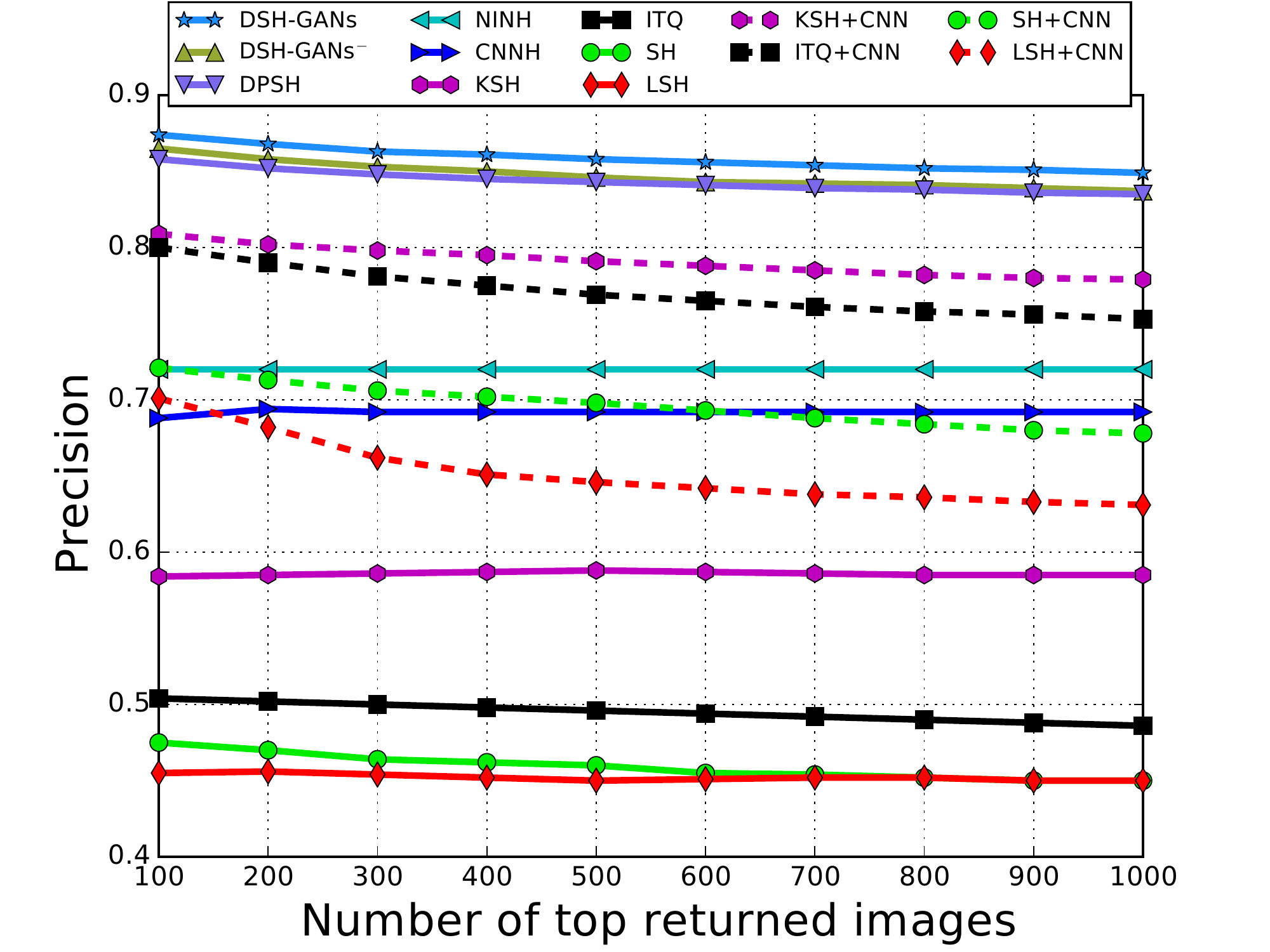}}
     \vspace{-0.08in}
   \caption{\small Comparisons with state-of-the-art approaches on NUS-WISE dataset. (a) Precision within Hamming radius 2 using hash lookup. (b) Precision-Recall curves with 48-bits. (c) precision curves with 48-bits w.r.t. different number of top returned samples. Better viewed in original color pdf file.}
   \label{fig:fig6}
  \vspace{-0.12in}
\end{figure*}

\begin{figure*}[!tb]
\centering {\includegraphics[width=0.98\textwidth]{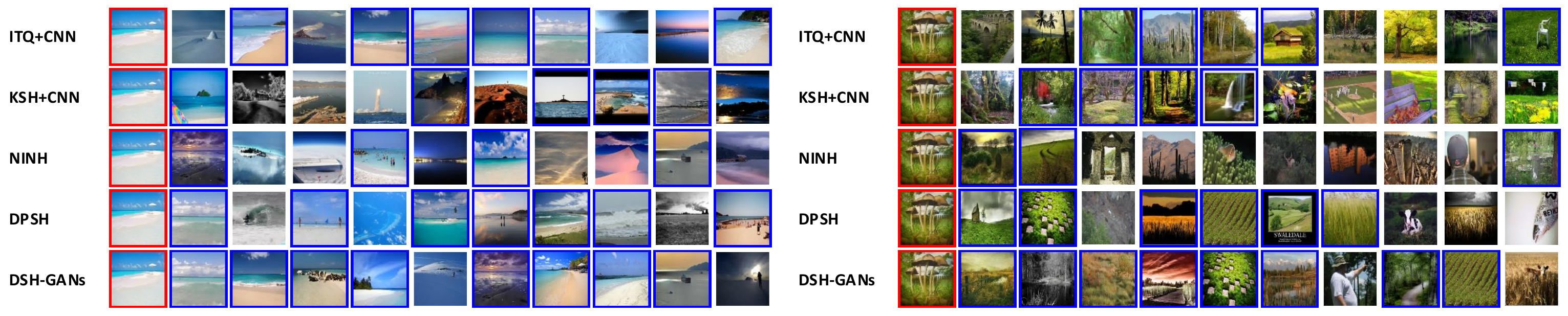}}
\caption{\small Examples showing the top 10 image retrieval results by different methods in response to two query images on NUS-WIDE dataset (better viewed in color). In each row, the first image with a red bounding box is the query image and the images whose annotations completely contain all the labels of the query image are regarded as excellent ones, which are enclosed in a blue bounding box.}
\label{fig:fig7}
\vspace{-0.15in}
\end{figure*}

\subsection{Results on NUS-WIDE Dataset}
The right half of Table \ref{tab:tab1} lists the MAP performance comparisons on NUS-WIDE dataset. Precision with Hamming radius 2 using hash lookup, precision-recall curves with 48-bits and precision curves with 48-bits w.r.t. different number of top returned samples is given in Figure \ref{fig:fig6:a}, \ref{fig:fig6:b} and \ref{fig:fig6:c}, respectively. DSH-GANs constantly exhibits better performance than other baselines across different performance metrics. Specifically, the MAP performance and precision with Hamming radius 2 using hash lookup of DSH-GANs achieve 86.3\% and 81.2\% with 48-bits, which make the improvements over the best competitor DPSH by 1.4\% and 2.7\%, respectively. This again verifies the effectiveness of generating synthetic and discriminable training data through GANs for hashing. Furthermore, DSH-GANs is benefited from utilizing semantic supervision and thus shows a relative increase of 1.1\%$\sim$1.9\% over DSH-GANs$^-$ in terms of MAP.

Figure \ref{fig:fig7} further showcases the top ten image search results by different methods in response to two query images. We can see that the proposed DSH-GANs method achieves the most satisfying results and retrieves eight ``excellent images" in the returned top ten images to each query image. It is worth noticing that ``excellent images" here refer to images whose annotations completely contain all the labels of the query image (e.g., ``water," ``clouds," ``ocean" and ``beach" of the first image example). As a result, the images retrieved by our DSH-GANs approach are more similar in semantics with the query image.

\begin{figure}[!tb]
   \centering
   \subfigure[CIFAR-10.]{
     \label{fig:fig8:a}
     \includegraphics[width=0.235\textwidth]{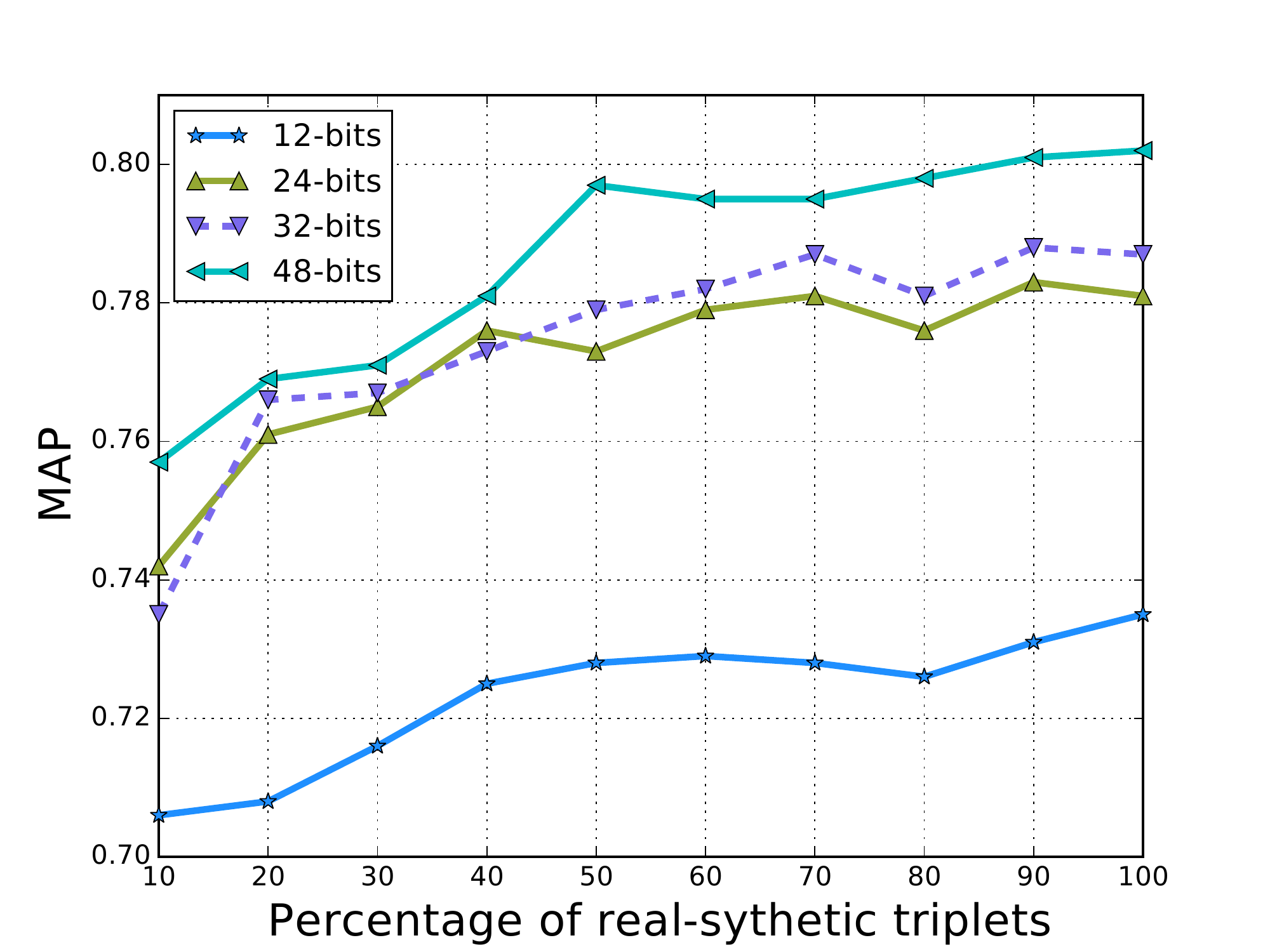}}
   \hspace{-4mm}
   \subfigure[NUS-WIDE.]{
     \label{fig:fig8:b}
     \includegraphics[width=0.235\textwidth]{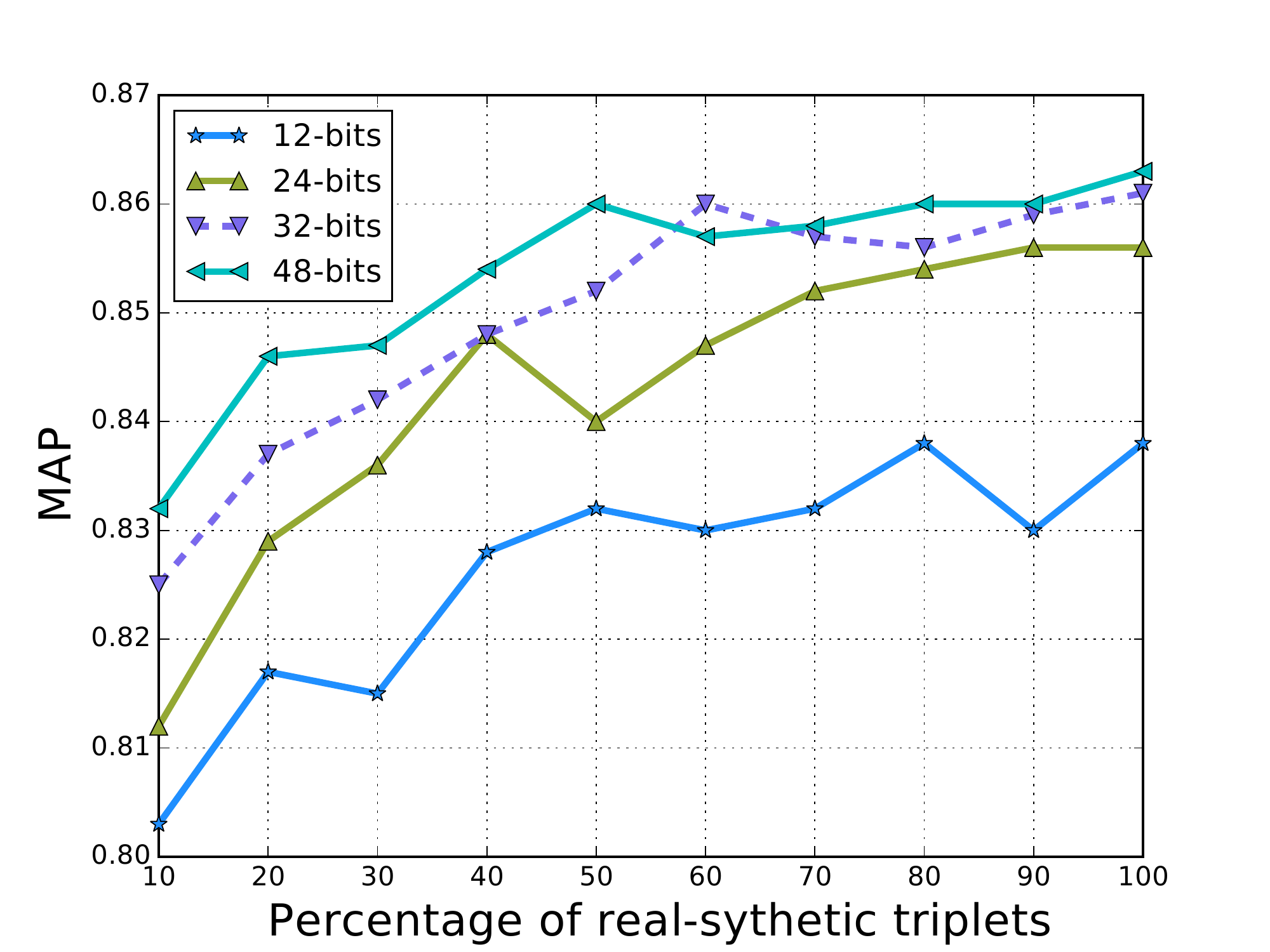}}
   \caption{\small MAP performance comparison with different percentage of synthetic data in training triplets.}
   \label{fig:fig8}
   \vspace{-0.12in}
\end{figure}

\begin{figure*}[!tb]
\centering {\includegraphics[width=0.95\textwidth]{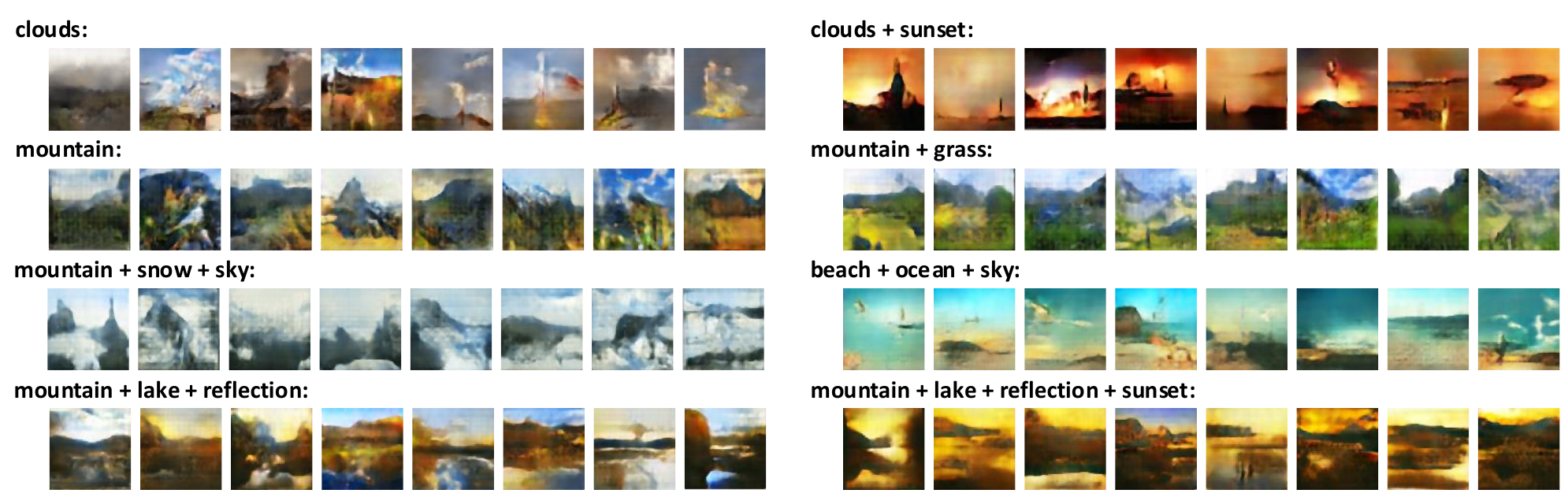}}
\caption{\small Visualization of synthetic image examples on NUS-WIDE dataset. All the image examples are generated with multiple labels. The images in the right half of each row are semantically related to the images in the left half.}
\label{fig:fig4}
\vspace{-0.12in}
\end{figure*}

\begin{figure}[!tb]
\centering {\includegraphics[width=0.5\textwidth]{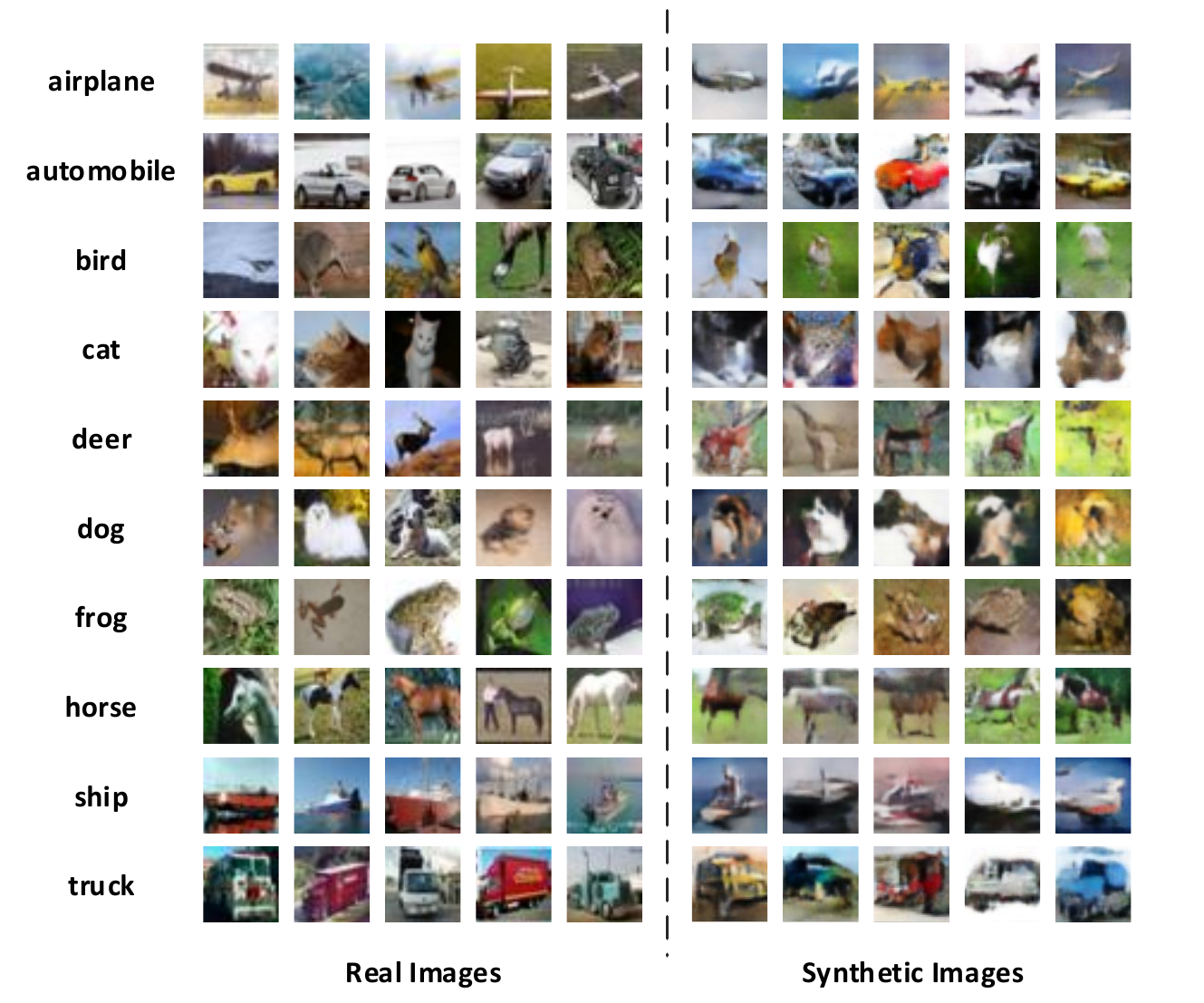}}
\caption{\small Visualization of image examples on CIFAR-10 dataset. Left half: images randomly selected from each class in the dataset; Right half: synthetic image examples for each class through our semi-supervised GANs.}
\label{fig:fig3}
 \vspace{-0.12in}
\end{figure}

\subsection{Comparison between Synthetic and Real Examples for Hashing}
In order to examine how performance is affected when exploiting synthetic examples in training triplets of different degree by DSH-GANs, we compare the MAP performances of using synthetic data with percentage ranging from 10\% to 100\%. In the previous experiments, the similar and dissimilar images in the training triplets are all synthetic images, which refers to 100\% in this analysis. We control the ratio between real and synthetic data in training by replacing part of synthetic images with real ones. Figure \ref{fig:fig8} shows the results on both CIFA-10 and NUS-WIDE datasets across different hash bits. The results are encouraging in the way that involving more synthetic data tends to achieve better performance. This empirically validates our proposal of generating synthetic data through semi-supervised GANs which additionally leverages largely unlabeled data, making the generated examples more discriminable to characterize the structure of the data.

\subsection{Visualization of Synthetic Images}
Figure \ref{fig:fig3} illustrates image examples on CIFAR-10 dataset, which are both randomly selected from each class in the dataset (left half) and generated for each class through our semi-supervised GANs (right half). In general, the generated images are plausible and semantically relevant to each class. Figure \ref{fig:fig4} further visualizes the synthetic image examples on NUS-WIDE dataset. The images in the right half of each row are semantically related to the images in the left half. Take the first row as an example, the images in the left half are generated with label ``clouds," while the images in the right half are synthesized with labels ``clouds" and ``sunset." All the images look real and the generated images in the right part could clearly manifest the semantics of ``sunset" and differentiate them from the images in the left part with only semantics of ``clouds."

\section{Conclusions}
We have presented a Deep Semantic Hashing with Generative Adversarial Networks (DSH-GA) architecture which explores semi-supervised GANs to generate synthetic training data for hashing. Particularly, a semi-supervised GANs is trained on both labeled and unlabeled data to produce compelling and discriminable examples conditioning on class labels. To verify our claim, we optimize the whole architecture of our hashing model by simultaneously distinguishing synthetic images from real ones and preserving not only relative similarity between images but also semantics of images. Experiments conducted on both CIFAR-10 and NUS-WIDE datasets validate our proposal and analysis. Performance improvements are clearly observed when comparing to other hashing techniques.

Our future works are as follows. First, as our architecture is a joint learning procedure, how the architecture performs on classification task will be further evaluated. Next, more in-depth studies of how to fuse the three streams in a principled way could be explored. Second, more advanced GANs (e.g., Stacked GANs) and CNN architectures (e.g., ResNet) will be investigated in our architecture.

\bibliographystyle{ACM-Reference-Format}
\bibliography{sigproc}

\end{document}